%% file: main.tex
\documentclass[10pt,twocolumn,letterpaper]{article}
\usepackage{tabularx}
\usepackage{float}
\usepackage{afterpage}
\usepackage{xspace}
\usepackage{comment}
\usepackage{cvpr}              


\definecolor{cvprblue}{rgb}{0.21,0.49,0.74}
\usepackage[pagebackref,breaklinks,colorlinks,allcolors=cvprblue]{hyperref}

\def\paperID{14} 
\def\confName{CVPR}
\def\confYear{2025}

\title{SmoothCache: A Universal Inference Acceleration Technique for Diffusion Transformers}

\author{
Joseph Liu\\
Roblox\\
{\tt\small josephliu@roblox.com}
\and
Joshua Geddes\\
Queen's University\\
{\tt\small j.geddes@queensu.ca}
\and
Ziyu Guo\\
Roblox\\
{\tt\small zguo@roblox.com}
\and
Haomiao Jiang\\
Roblox\\
{\tt\small haomiaojiang@roblox.com}
\and
Mahesh Kumar Nandwana\\
Roblox\\
{\tt\small mnandwana@roblox.com}
}

\begin{document}

\twocolumn[{%
\renewcommand\twocolumn[1][]{#1}%
\maketitle
\begin{center}
    \vspace{-10mm}
    \centering
    \captionsetup{type=figure}
    \includegraphics[width=1\textwidth]{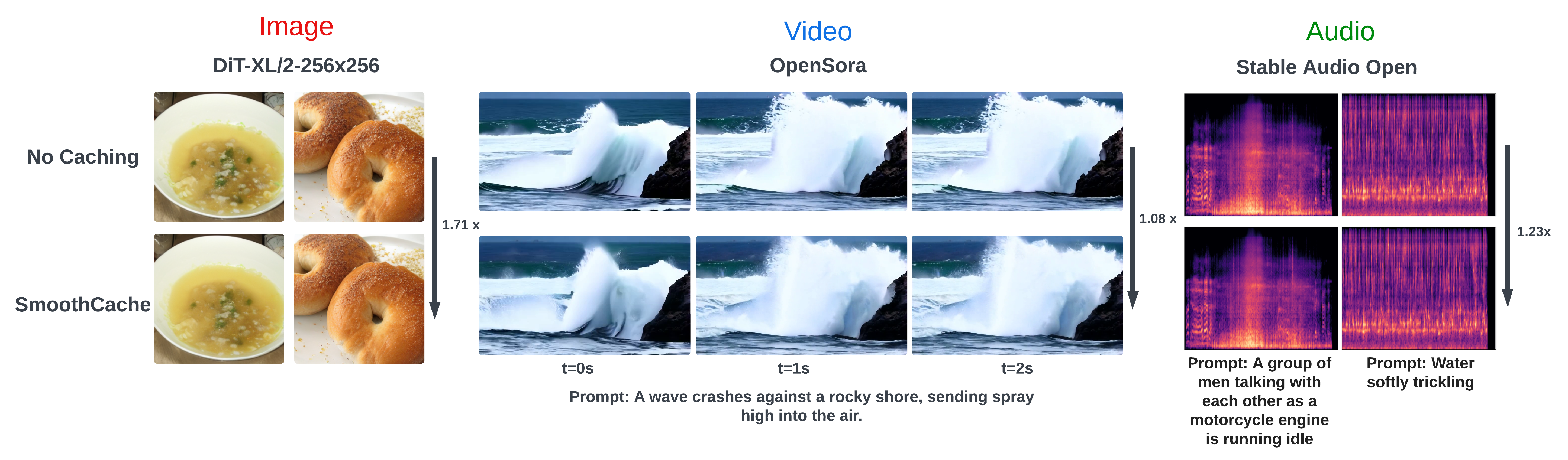}
    \vspace{-9mm}
    \captionof{figure}{Accelerating Diffusion Transformer inference across multiple modalities with 50 DDIM Steps on DiT-XL-256x256, 100 DPMSolver++(3M) SDE steps for a 10s audio sample (spectrogram shown) on Stable Audio Open,  30 Rectified Flow steps on Open-Sora 480p 2s videos.}
    \label{fig:teaser_figure}
    \vspace{-5mm}
\end{center}%
}]
\input{sec/0_abstract}

\input{sec/1_intro}
\input{sec/2_main_content}
\input{sec/3_conclusion}
{
    \small
    \bibliographystyle{ieeenat_fullname}
    \bibliography{main}
}

\input{sec/X_suppl}

\end{document}

%% file: sec/0_abstract.tex
\begin{abstract}
Diffusion Transformers (DiT) have emerged as powerful generative models for various tasks, including image, video, and speech synthesis. However, their inference process remains computationally expensive due to the repeated evaluation of resource-intensive attention and feed-forward modules. To address this, we introduce SmoothCache\footnote{Code can be found at https://github.com/Roblox/SmoothCache}, a model-agnostic inference acceleration technique for DiT architectures. SmoothCache leverages the observed high similarity between layer outputs across adjacent diffusion timesteps. By analyzing layer-wise representation errors from a small calibration set, SmoothCache adaptively caches and reuses key features during inference. Our experiments demonstrate that SmoothCache achieves 8\% to 71\% speed up while maintaining or even improving generation quality across diverse modalities. We showcase its effectiveness on DiT-XL for image generation, Open-Sora for text-to-video, and Stable Audio Open for text-to-audio, highlighting its potential to enable real-time applications and broaden the accessibility of powerful DiT models.
\end{abstract}

%% file: sec/1_intro.tex

\section{Introduction}
\label{sec:intro}

In the rapidly evolving landscape of generative modeling, Diffusion models~\cite{sohldickstein2015deepunsupervisedlearningusing, ho2020denoisingdiffusionprobabilisticmodels} have emerged as a pivotal force, offering unparalleled capabilities for creating rich and diverse content. Diffusion transformers (DiT)~\cite{peebles2023scalablediffusionmodelstransformers,vaswani2023attentionneed}, in particular, harnessing the scalable architecture of transformers, demonstrated significant advancements in various domains, including but not limited to the creation of images~\cite{chen2024pixartalpha}, audio~\cite{evans2024stable, le2023voiceboxtextguidedmultilingualuniversal, liu2024vitttsvisualtexttospeechscalable}, video~\cite{opensora, pku_yuan_lab_and_tuzhan_ai_etc_2024_10948109, ma2024lattelatentdiffusiontransformer}, and 3D models~\cite{mo2023dit3dexploringplaindiffusion}.


The central challenge limiting the broader adoption of DiT's is the computational intensity of their inference process.  The cost of running the entire diffusion pipeline lies primarily in the denoising steps, and improving denoising efficiency directly contributes to a faster diffusion pipeline.  To address this, inference optimizations have been researched from two directions: (1) reducing the number of sampling steps,  and (2) lowering the inference cost per step.


The use of advanced solvers~\cite{lu2023dpmsolverfastsolverguided,liu2022flow} have been proposed  to reduce the number of timesteps required for sampling, thereby accelerating inference.  Techniques such as knowledge distillation~\cite{salimans2022progressivedistillationfastsampling}, architecture optimizations~\cite{li2023snapfusiontexttoimagediffusionmodel}, pruning~\cite{fang2023structuralpruningdiffusionmodels}, and quantization~\cite{shang2023posttrainingquantizationdiffusionmodels,he2023ptqdaccurateposttrainingquantization} aim to reduce the computational complexity of individual denoising steps.



While these methods have shown promise, there remains a need for techniques that can further accelerate DiT inference \emph{across diverse modalities} without compromising generation quality. Caching~\cite{ma2023deepcacheacceleratingdiffusionmodels,wimbauer2024cachecanacceleratingdiffusion,liu2024fasterdiffusiontemporalattention,agarwal2023approximatecachingefficientlyserving} has emerged as a potential solution by exploiting the inherent redundancy in the diffusion process. Previous and concurrent work on DiT caching has investigated uniform scheduling~\cite{selvaraju2024forafastforwardcachingdiffusion}, training a caching pattern with a dataset~\cite{ma2024learningtocacheacceleratingdiffusiontransformer}, and exploiting qualitative characteristics about a candidate model to create a caching pattern~\cite{ chen2024deltadittrainingfreeaccelerationmethod,zhao2024realtimevideogenerationpyramid}. However, these caching techniques are limited by either overly simplistic or model-specific strategies or reliance on costly retraining procedures.


To overcome these limitations, we introduce SmoothCache, a simple and universal caching scheme capable of speeding up a diverse spectrum of Diffusion Transformer models. SmoothCache leverages the observation that layer outputs from adjacent timesteps in any DiT-based model exhibits high cosine similarities~\cite{selvaraju2024forafastforwardcachingdiffusion, ma2024learningtocacheacceleratingdiffusiontransformer}, suggesting potential computational redundancies. By carefully leveraging layer-wise representation errors from a small calibration set, SmoothCache adaptively determines the optimal caching intensity at different stages of the denoising process. This allows for the reuse of key features during inference without significantly impacting generation quality. SmoothCache is designed with generality in mind, and can be applied to any DiT architecture without model-specific assumptions or training while still achieving performance gains over uniform caching.

We show that SmoothCache with general training-free caching scheme is able to speedup the performance across image, video, audio domains to match or exceed SOTA caching scheme for each dedicated domain. We demonstrate that SmoothCache is also compatible with various common solvers in diffusion transformers.

\subsection{Related Work}

\label{related_work}

Diffusion models \cite{sohldickstein2015deepunsupervisedlearningusing, ho2020denoisingdiffusionprobabilisticmodels} have emerged as a powerful class of generative models, capable of producing high-quality samples across various domains such as images, audio, and video. Initally, U-Net architectures were favored for this denoising step due to their strong ability to capture both global and local information in the latent space, crucial for reconstructing fine details in the data \cite{rombach2022highresolutionimagesynthesislatent, saharia2022photorealistictexttoimagediffusionmodels}. 

However, the inherent limitations of U-Nets, particularly their struggle to scale effectively to high-dimensional data and long sequences, led to the exploration of alternative architectures. The Diffusion Transformer (DiT) \cite{peebles2023scalablediffusionmodelstransformers} was proposed as a solution to U-Nets to exploit the scalability of the transformer architecture \cite{vaswani2023attentionneed}. Transformers, renowned for their ability to handle long-range dependencies and their inherent scalability, have proven highly effective in various sequence-based tasks. DiT successfully adapted these advantages to the diffusion framework, demonstrating strong performance in generating not only images but also extending to diverse domains like speech \cite{evans2024stable, le2023voiceboxtextguidedmultilingualuniversal, liu2024vitttsvisualtexttospeechscalable}, video \cite{opensora, pku_yuan_lab_and_tuzhan_ai_etc_2024_10948109, ma2024lattelatentdiffusiontransformer}, and 3D generation \cite{mo2023dit3dexploringplaindiffusion}.

\noindent \textbf{Efficient Diffusion Models. }
Diffusion models require many function evaluations to iteratively remove noise from data, and thus quickly scale in computational costs. Many previous works have focused on reducing the number of diffusion steps required to achieve high quality samples by exploring accelerated noise schedulers and fast ODE solvers~\cite{song2022denoisingdiffusionimplicitmodels,lu2022dpmsolverfastodesolver,lu2023dpmsolverfastsolverguided,liu2022flow}. However, not all tasks might be suitable for faster solvers with less steps, as they might rely on inversion for tasks like image-editing~\cite{mokady2023null}, and while work has been done to extend this to other solvers~\cite{hong2023exactinversiondpmsolvers,wang2024tamingrectifiedflowinversion}, future solvers will face similar issues with adoption.
Traditional acceleration methods such as such as quantization \cite{shang2023posttrainingquantizationdiffusionmodels,li2023snapfusiontexttoimagediffusionmodel,he2023ptqdaccurateposttrainingquantization} and compression/distillation techniques \cite{li2023snapfusiontexttoimagediffusionmodel,salimans2022progressivedistillationfastsampling,fang2023structuralpruningdiffusionmodels} could improve model performance, but these techniques often require extensive retraining, or specific architectures and parameterization around a specific task in order to realize efficiency gains.

\noindent \textbf{Diffusion Model Caching.} Caching has emerged as a promising technique for accelerating DiT inference by exploiting computational redundancies in the denoising process. For U-Net based architectures, DeepCache \cite{ma2023deepcacheacceleratingdiffusionmodels} leverages cross-timestep similarities by caching and reusing up sampling feature maps, while other methods use focus on caching blocks of layers \cite{wimbauer2024cachecanacceleratingdiffusion}, cross-attention modules \cite{liu2024fasterdiffusiontemporalattention}, or intermediate noise states \cite{agarwal2023approximatecachingefficientlyserving}. 
However, in addition to exploiting specific properties of the U-Net architecture, these models only perform these analyses on image diffusion models, leveraging biases and assumptions in those tasks that might not necessarily generalize to other modalities.
For Diffusion Transformers, further research has been conducted to apply caching to image~\cite{ma2024learningtocacheacceleratingdiffusiontransformer, chen2024deltadittrainingfreeaccelerationmethod, selvaraju2024forafastforwardcachingdiffusion} and video~\cite{zhao2024realtimevideogenerationpyramid} DiT diffusion models. Again, these methods only study how their methods accelerate diffusion inference in their specific modalities, with very little attempt to show that these methods generalize beyond their modality of interest. FORA~\cite{selvaraju2024forafastforwardcachingdiffusion} for example might have extensive acceleration gains in image diffusion, but our preliminary investigations shows that it does not work on Audio or Video diffusion tasks. Likewise, Pyramid-Attention Broadcast~\cite{zhao2024realtimevideogenerationpyramid} leverages specific qualities of video diffusion, such as its cache-able cross-attention layer in order to accelerate diffusion, on top of other GPU parallelization tricks, but Fig.~\ref{fig:component_error_curves} shows this assumption breaks in other modalities.
Moreover, not all caching methods are post-training. Training-based caching methods like Learning-to-Cache~\cite{ma2024learningtocacheacceleratingdiffusiontransformer} that require further training for specific diffusion step configurations, which is not viable in cases where access to source training data might not be available, such as in open source models trained on proprietary (non-public) data.

SmoothCache attempts to address these limitations by introducing a training-free caching approach, that makes no underlying assumption about the specific task or modality at hand, outperforming uniform schedules and even matching or beating handcrafted caching techniques reported in literature. We also present multi-modal results to show that this technique does generalize across modalities, and how it adapts to various model, sampling, and solver configurations with just one calibration inference pass and a single hyper-parameter $\alpha$.

\afterpage{
    \begin{figure*}[h]
        \centering
        \includegraphics[width=0.9\textwidth]{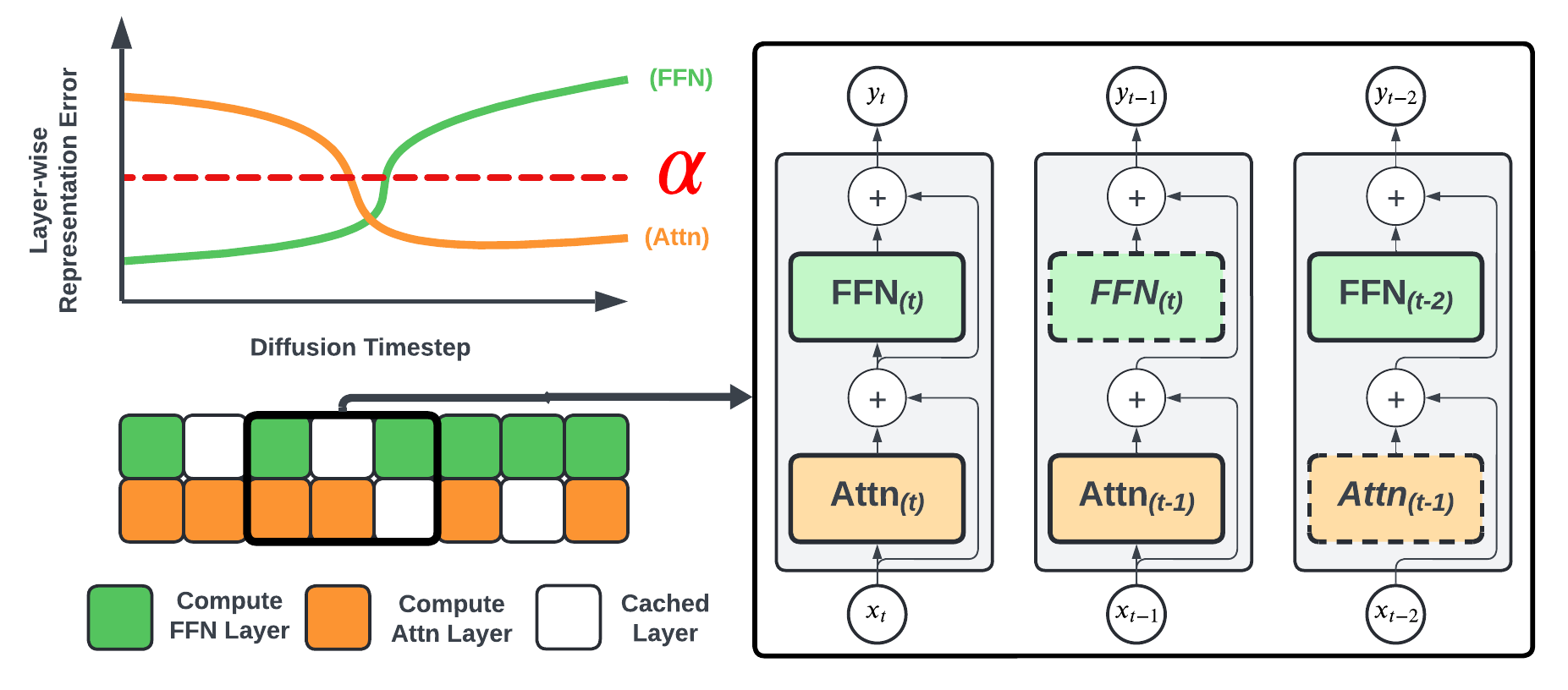}
        \vspace{-4mm}
        \caption{\textbf{Illustration of SmoothCache.} When the layer representation loss obtained from the calibration pass is below some threshold $\alpha$, the corresponding layer is cached and used in place of the same computation on a future timestep. The figure on the left shows how the layer representation error impacts whether certain layers are eligible for caching. The error of the attention (attn) layer is higher in earlier timesteps, so our schedule caches the later timesteps accordingly. The figure on the right shows the application of the caching schedule to the DiT-XL architecture. The output of the attn layer at time $t-1$ is cached and re-used in place of computing FFN $t-2$, since the corresponding error is below $\alpha$. This cached output is introduced in the model using the properties of the residual connection.}
        \label{fig:smoothcache}
        
\vspace{-4mm}
    \end{figure*}
}

%% file: sec/2_main_content.tex
\section{Method}
\label{methods}

\begin{figure}[t]
    \centering
    \includegraphics[width=0.45\textwidth]{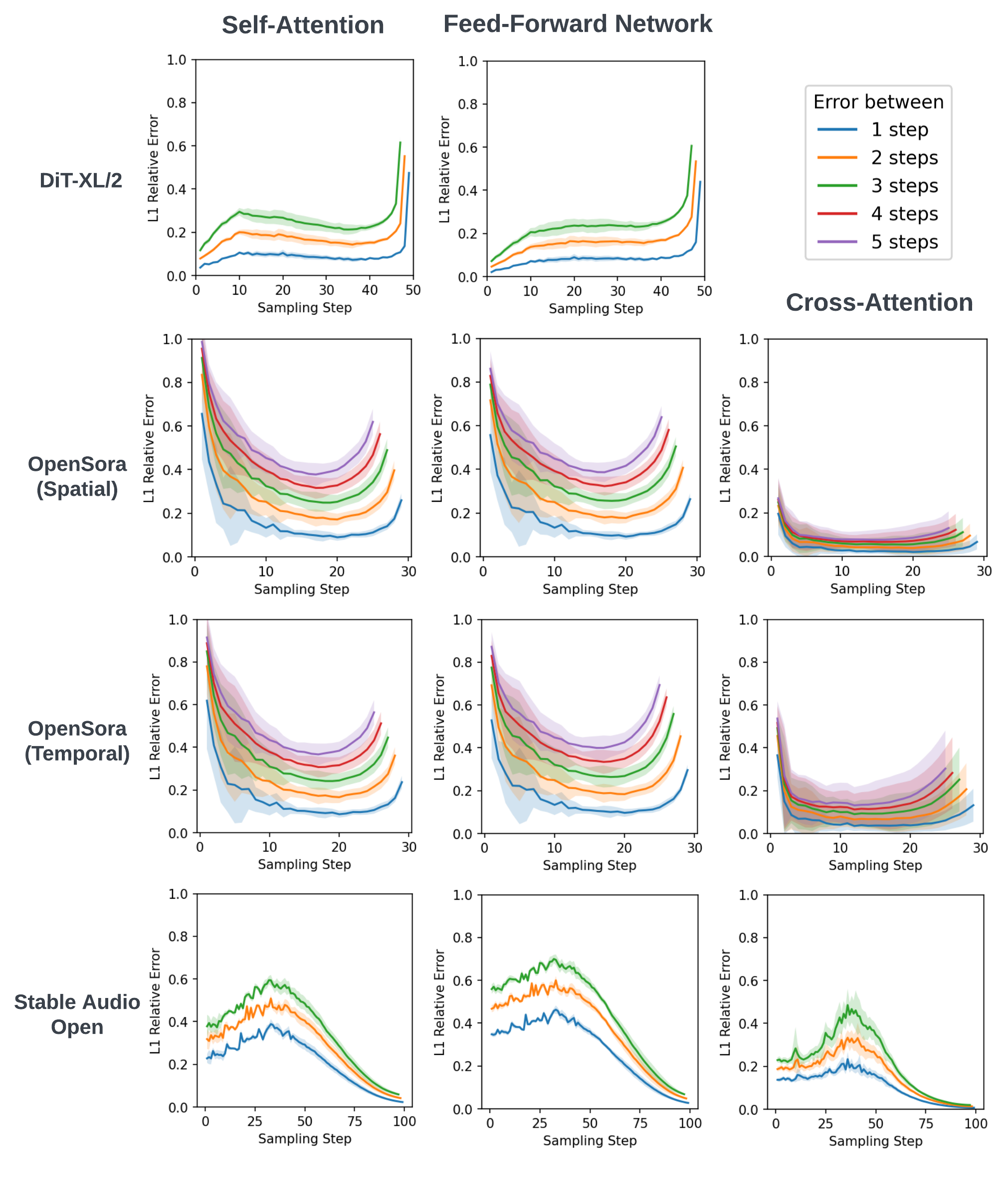}
    \vspace{-2mm}
    \caption{\textbf{L1 Relative Error Curves of different architecture components}. Curves are plotted with 95\% confidence intervals from 10 calibration samples from all components explored in this paper and scaled to the same y-axis range. Note that OpenSora has distinct spatial and temporal diffusion blocks.}
    \label{fig:component_error_curves}
    
\vspace{-6mm}
\end{figure}
\label{methods}
In this section, we describe the preliminary setup, base assumptions, observations and math fomulation of the proposed SmoothCache method.

\subsection{Preliminaries}

The diffusion process transforms data $\mathbf{x}_0 \sim q(\mathbf{x}_0)$ by adding Gaussian noise over $T$ steps, producing a sequence $\mathbf{x}_1, \mathbf{x}_2, \dots, \mathbf{x}_T$. This is modeled as a Markov chain, where each step follows:

\begin{equation}
q(\mathbf{x}_t | \mathbf{x}_{t-1}) = \mathcal{N}(\mathbf{x}_t; \sqrt{1 - \beta_t} \mathbf{x}_{t-1}, \beta_t \mathbf{I}),
\end{equation}

where $\beta_t$ is the noise variance at timestep $t$ and $\mathbf{x}_T \sim \mathcal{N}(0, \mathbf{I})$ due to the cumulative effect of the noise added during the diffusion steps. The goal of the reverse process is to recover the original data $\mathbf{x}_0$ from the noisy sample $\mathbf{x}_T$. This is learned via a neural network that parameterizes the reverse transition:

\begin{equation}
p_{\theta}(\mathbf{x}_{t-1} | \mathbf{x}_t) = \mathcal{N}(\mathbf{x}_{t-1}; \mu_{\theta}(\mathbf{x}_t, t), \Sigma_{\theta}(t)),
\end{equation}

where $\mu_{\theta}(\mathbf{x}_t, t)$ is the predicted mean and $\Sigma_{\theta}(t)$ is the variance, both learned by the model. The full generative process is defined as:

\begin{equation}
p_{\theta}(\mathbf{x}_{0:T}) = p(\mathbf{x}_T) \prod_{t=1}^{T} p_{\theta}(\mathbf{x}_{t-1} | \mathbf{x}_t).
\end{equation}


The generative model is trained by minimizing the variational bound on the negative log likelihood, learning to denoise $\mathbf{x}_t$ at each timestep and ultimately generate samples from the learned distribution. Traditionally, $p_{\theta}(\mathbf{x}_{t-1} | \mathbf{x}_t)$ has been approximated with U-Net style model architectures, but recent works have begun to use Diffusion Transformer (DiT) architectures, which have shown to scale better, especially for more complex tasks such as video generation. DiT architectures consist of repeated blocks containing the Self-attention, Cross-attention, and Feed-forward layers in the traditional Transformer, which are usually the computational bottleneck for both model training and inference.

A key property of diffusion models, which has driven the development of caching techniques, is the high cosine similarity between layer outputs at adjacent timesteps. This pattern of similarity is observed across a variety of generative models and solvers, spanning different modalities such as image, video, and speech diffusion. This high similarity suggests that there are computational redundancies within the diffusion process that can be leveraged to improve efficiency.



\subsection{SmoothCache}

The objective of SmoothCache is to provide a training-free, model-agnostic strategy of caching and reusing layer outputs such that minimal error is introduced. However, we believe a single optimal static scheme that caches cross-timestep layer output may not exist across different model architectures, solvers, and modalities. For example, we examine the average representation error between layer outputs of consecutive time steps as shown in Fig.~\ref{fig:component_error_curves}. For layer output $L$ at timestep $t$ the representation error $k$ timesteps behind is defined as $E(L_{t}, L_{t+k})=\frac{\| L_{t} - L_{t+k} \|_1}{\| L_{t} \|_1}$.


We observe that layers generally exhibit higher differences in later time steps in the DiT-XL label-to-image model, while layers in the Open-Sora text-to-video model are more sensitive in the first and last diffusion time steps. This discovery emphasizes a need to apply the similarity principle in a generalizable technique, such that different models benefit differently based on error curves.

\begin{figure}[b]
\vspace{-4mm}
    \centering
    \includegraphics[width=0.9\columnwidth]{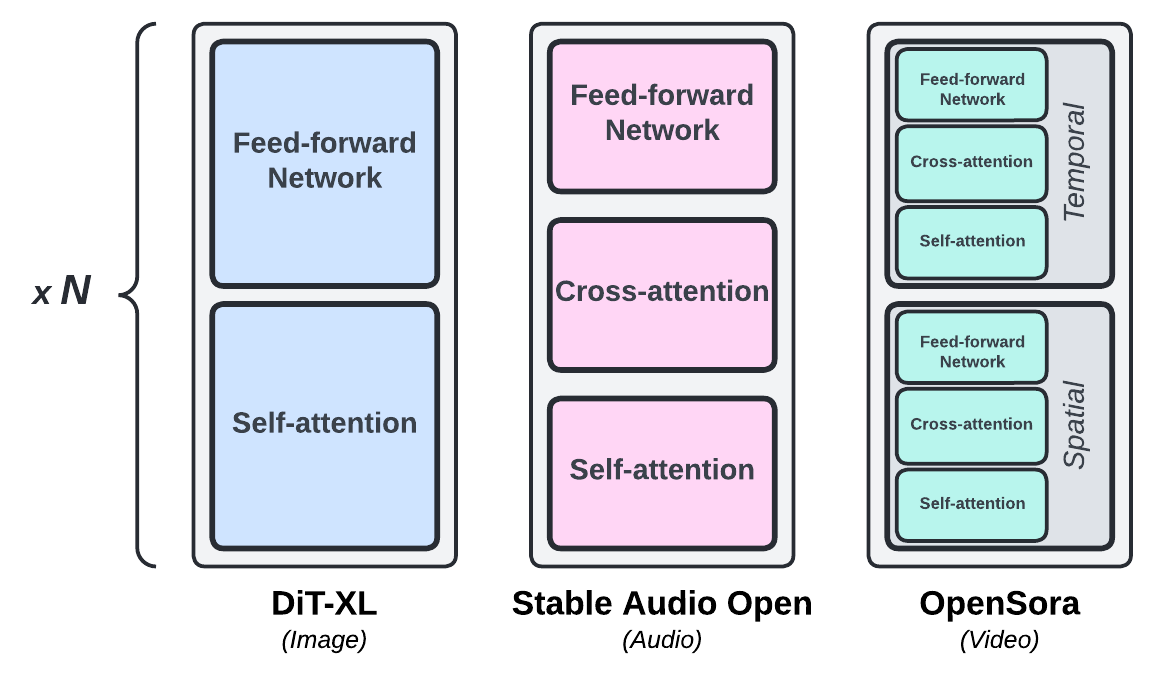}
    \vspace{-5mm}
    \caption{\textbf{SmoothCache-Eligible Layers of candidate models}. This visualization highlights the targeted layers that precede residual connections in a DiT block for each architecture. Each model contains $N$ DiT blocks. In the original DiT-XL model, Self-attention and Feed-forward layers are cached. In the Stable Audio Open model, Self-attention, Cross-attention, and Feed-forward layers are cached. In the Open Sora model, Self-attention, Cross-attention, and Feed-forward layers across both the temporal and spatial partitions of the DiT block.}
    \label{fig:layers}
\end{figure}

\begin{figure}[t]
    \centering
    \includegraphics[width=0.9\columnwidth]{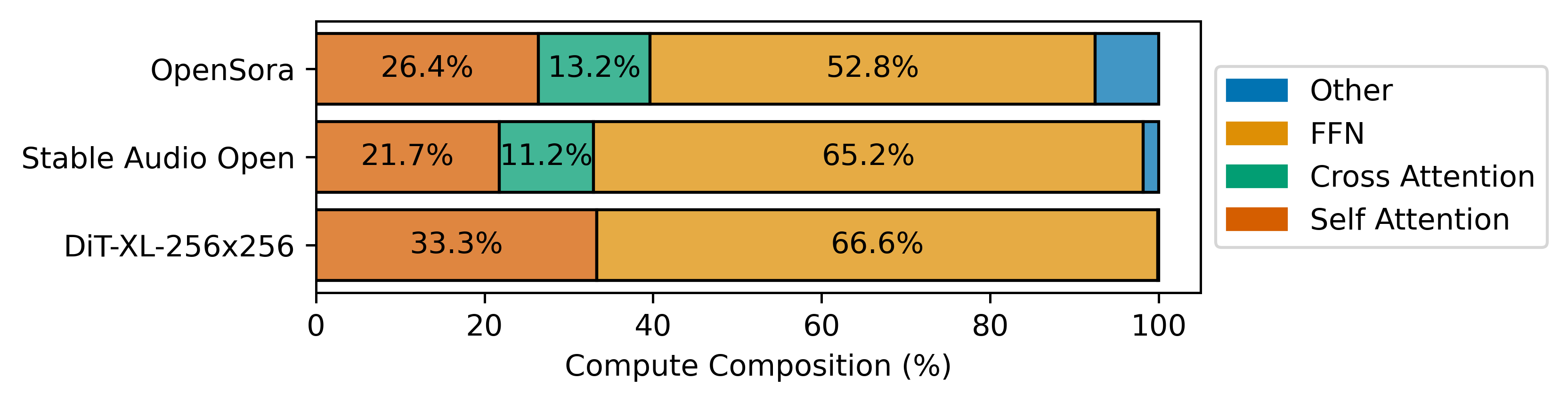}
    \vspace{-1mm}
    \caption{\textbf{Layer Compute Composition of candidate models}. These are computed from the MACs of the default model configurations without SmoothCache applied. Note that in all candidate models, SmoothCache eligible layers comprise at least 90\% of compute time.}
    \label{fig:layers_compute}
\vspace{-4mm}
\end{figure}

Let $L_{t}$ represent the output of some layer at timestep $t$, and let $L_{t-k}$ represent the output of the same layer at some future diffusion timestep $t-k$. By the cross-timestep layer similarity observation, $L_{t-k} \sim L_{t}$. Thus instead of computing $L_{t-k}$ during the diffusion process, we can approximate the function by using the previously computed $L_{t}$. Any time the layer is computed, our method stores $L_{t}$ in a cache that can be accessed and used in place of future layer computations. We apply caching to the aforementioned computational bottlenecks at the output that precedes a residual connection, shown in Fig.~\ref{fig:smoothcache}.  This includes Self-attention and Feed-forward layers in the DiT-XL model, Self-attention, Cross-attention, and Feed-forward layers in the StableAudio model, and Self-attention, Cross-attention, and Feed-forward layers in both the spatial and temporal blocks in the OpenSora model as shown in Fig.~\ref{fig:layers}. Fig.~\ref{fig:layers_compute} shows that these components comprise of nearly all of the compute that occurs during generation. We also note that the compute distribution varies from model to model.

\label{methods}

\label{methods}

To determine whether to use a previously cached output, we define the following problem. Let $t$ represent the current timestep, $t+k$ represent the timestep when the cache was previously filled, and $i_j$ represent the $j$th layer of type $i$, where $i \in S = \{attn, f\!f\!n, ...\}$ depending on the model architecture. Our method is guided by the key hypothesis that caching is effective if the loss between the computed and cached outputs $\mathcal{L}(L_{i_j,t}, L_{i_j,t+k})$, is bounded by some layer-dependent hyper parameter $\alpha_{i_j} > 0$.

Computing $\mathcal{L}(L_{i_j,t}, L_{i_j,t+k})$, is not possible without evaluating $L_{i_j,t}$, which removes any benefit from skipping the layer computation. For a given DiT architecture, we remarkably observe that the difference in layer representation error for two different samples is within a negligibly small threshold, as shown the 95\% confidence interval of the plots generated with 10 calibration samples in Fig.~\ref{fig:component_error_curves}. This finding suggests that the error curve for a specific model input $\mathcal{L}(L_{i_j,t}, L_{i_j,t+k})$ can be closely approximated by the average error curve for an adequately large set of calibration inputs. In other words, if $\tilde{L}_{i_j,t}$ represents the calibration output for layer $L_{i_j,t}$, then $\mathcal{L}(L_{i_j,t}, L_{i_j,t+k}) \sim \mathcal{L}(\tilde{L}_{i_j,t}, \tilde{L}_{i_j,t+k})$.

A hyper parameter search to find all $\alpha_{i_j}$ has a exponential search space based on the number of layers and is significantly costly. In order to simplify the caching problem, we define a single hyper parameter $\alpha > 0$ to guide caching for all layers. We define $\mathcal{L}$ as the average L1 relative error of all $N$ layers of type $i$, which is selected in order to compare true representation errors between layers for all types $i$ and and positional depths $j$ in the network. Additionally, we recognize caching specific layers can introduce errors in future layers of the same type in the network. For example, if we approximate $L_{i_{j-1},t}$ with $L_{i_{j-1},t+k}$, this may introduce noise such that the calibration error $\mathcal{L}(\tilde{L}_{i_j,t}, \tilde{L}_{i_j,t+k})$ no longer correctly approximates the true error $\mathcal{L}(L_{i_j,t}, L_{i_j,t+k})$, which leads to poor caching decisions. In order to mitigate the cascading impact of caching layers, we group caching decisions for all layers of type $i$, such that all $j$ layers in $L_{i,t+k}$ approximate $L_{i,t}$. Thus, cached output $L_{i_{j},t+k}$ is used in place of computing $L_{i_{j},t}$ when

\vspace{-4mm}
\begin{equation}
\label{eq:loss}
\mathcal{L}(L_{i_j,t}, L_{i_j,t+k}) \sim \frac{1}{N}\sum_{j=1}^{N}\frac{\| \tilde{L}_{i_j,t} - \tilde{L}_{i_j,t+k} \|_1}{\| \tilde{L}_{i_j,t} \|_1}
< \alpha
\end{equation}
Using cached outputs is computationally inexpensive, and allows for significant speedup of the diffusion inference process. As Eq.~\ref{eq:loss} makes no prior assumptions about the specific properties of the particular diffusion process being cached, it can be applied across multiple architectures and modalities, only requiring a brief linear search to find an optimal $\alpha$ hyperparameter for each sampling configuration. Additionally, because caching decisions are only dependent on calibration error, they do not change at model runtime. This ensures compatibility with existing graph compilation optimizations.

\section{Experiments}
\label{experiments}

\subsection{Experiment Setup}

\textbf{Models, Datasets, and Solvers} 
In order to demonstrate the effectiveness of our technique across DiT architectures and prove the model-agnostic claim, we evaluate SmoothCache on multiple candidate diffusion models across a variety of modalities, across different numbers of diffusion steps.

\textit{Text to Image Generation} Firstly, we select the original DiT-XL-256$\times$256 label to image diffusion model, using the original released model weights~\footnote{https://github.com/facebookresearch/DiT}. Since the DiT family of models is trained on the ImageNet1k dataset, we generate 50,000 256$\times$256 images for evaluation. We test SmoothCache with the DDIM solver, and use classifier free guidance with a scale of 1.5.

\textit{Text to Video Generation} Secondly, we select the OpenSORA text to video model. We use the pre-trained OpenSORA v1.2 model, and evaluate its performance using VBench evaluation protocol using videos generated from the 946 prompts from the VBench prompt suite on 2 second 480p videos with a 9:16 aspect ratio. We use flash attention and the bfloat16 datatype for inference. We test SmoothCache using the Rectified Flow~\cite{liu2022flow} solver for 30 sampling steps across 1000 Euler steps, using the default CFG scale of 7.0. 

\textit{Text to Audio Generation} Lastly, we select the Stable Audio Open text to audio model. We follow the same inference protocol as described in the original paper~\cite{evans2024stable}. We run inference Stable Audio Open with DPM-Solver++ (3M) SDE for 100 steps with classifier free guidance scale of 7.0.


\textbf{Evaluation Metrics}
We use a variety of standard metrics in each domain to demonstrate the efficiency to quality trade off. We record the Multiply-Accumulate Operations (MACs) and latency of the full diffusion process. We derive the acceleration ratio from baseline sampling latencies. To measure generation quality, we evaluate the candidate models on common metrics in the corresponding domain. 
For DiT-XL-256$\times$256, we generate 50,000 images with the specified configuration and report the FID, IS, and sFID. 
For OpenSORA, we evaluate the performance of the model using VBench~\cite{huang2023vbench}, Learned Perceptual Image Patch Similarity (LPIPS)~\cite{zhang2018unreasonable}, Peak Signal-to-Noise Ratio (PSNR) and Structural Similarity
Index Measure (SSIM)~\cite{ssim2002}, and generate 946 videos based on the VBench suite prompts. LPIPS, PSNR and SSIM are computed relative to the non-cached videos. We report the final scaled VBench score. 
Lastly, for Stable Audio Open, we use the exact same evaluation protocol described in the Stable Audio Open technical report \cite{evans2024stable} using the same evaluation code\footnote{https://github.com/Stability-AI/stable-audio-metrics}, reporting CLAP, \texorpdfstring{FD\textsubscript{OpenL3}}{FDopenl3} and \texorpdfstring{KL\textsubscript{PaSST}}{KLpasst} metrics for AudioCaps~\cite{audiocaps}, MusicCaps without singing prompts~\cite{musiccaps_gostinelli2023musiclmgeneratingmusictext}, and Song Describer without singing prompts~\cite{manco2023songdescriber}. All speed results are measured on a single H100-80G GPU, and averaged across 50 runs.

\textbf{Implementation Details}
As mentioned previously, we apply caching to layers that precede residual connections. In the DiT-XL-256$\times$256 model, this includes the Self-attention and Feed-forward modules in the DiT block. In OpenSora V1.2, we cache the temporal Self-attention, Cross-attention and Feed-forward modules as well as their equivalent spatial variants (for a total of 6 types of modules). 
In Stable Audio Open, we cache the Self-attention, Cross-attention and Feed-forward modules. For each architecture, solver, and preset number of diffusion steps, we obtain the representation errors $L(F_{i_j,t}, F_{i_j,t+k})$ by computing differences in layer outputs for some random samples. 
For all modalities, we use 10 samples for calibration. Ablations show that the choice of the number of calibration samples does not matter much, something that can be observed in Fig.~\ref{fig:component_error_curves}.
We also choose which samples to generate the error curves based on whether we did conditional generation during calibration or not, which we only do for OpenSora and Stable Audio Open.
We fix $k \in \{1, 2, 3\}$ for DiT-XL-256$\times$256 and Stable Audio Open as we determine the layer representation error to grow too large past a difference of 4 timesteps. In OpenSORA, $k$ goes up to 5 in particular due to relatively low error in the cross-attention components of the network.
For DiT, we calibrate on samples generated unconditionally using the null prompt. We calibrate OpenSora on conditionally generated 480p 2s videos with randomly sampled prompts from VidProM~\cite{wang2024vidprom}. 
For Stable Audio Open, we calibrate using randomly sampled prompts from the AudioCaps validation set. 

\subsection{Results}
We present the results of SmoothCache on DiT-XL-256$\times$256, OpenSora, and Stable Audio Open in Tables~\ref{tab:dit_table},~\ref{tab:opensora_table}, and~\ref{tab:stableaudio_table} respectively. In order to provide a fair comparison between acceleration and quality trade offs, we compare the default solver used in the model, such as DDIM or DPM++, with a SmoothCache implementation with specified hyper parameters. We find two configurations with identical acceleration ratios show SmoothCache to perform better on various quality metrics, indicating superior performance over the typical acceleration/quality tradeoff. For all results, we run 5 trials and report the mean and standard deviation for each metric.

\begin{table}[t]
\centering
\vspace{-1mm}
\caption{Results For DiT-XL-256$\times$256 on using DDIM Sampling, sorted by TMACs. Note that L2C is not training free.}
\fontsize{6}{7}\selectfont
\setlength{\tabcolsep}{3pt}
\renewcommand{\arraystretch}{1.2}
\vspace{-1mm}
\begin{tabularx}{1\columnwidth}{c|c|c c c | c c}
\toprule
Schedule  & Steps & FID ($\downarrow$) & sFID ($\downarrow$) & IS ($\uparrow$) & TMACs  & Latency (s) \\
\midrule
 L2C  & 50 & 2.27 $\scriptscriptstyle$$\pm$$\scriptscriptstyle$ 0.04 & 4.23 $\scriptscriptstyle$$\pm$$\scriptscriptstyle$ 0.02 & 245.8 $\scriptscriptstyle$$\pm$$\scriptscriptstyle$ 0.7 & 278.71 & 6.85\\
\midrule
 No Cache  & 50 & 2.28 $\scriptscriptstyle$$\pm${$\scriptscriptstyle$0.03} & 4.30 $\scriptscriptstyle$$\pm$$\scriptscriptstyle$0.02 & 241.6 $\scriptscriptstyle$$\pm$$\scriptscriptstyle$ 1.1 & 365.59 & 8.34\\
 Ours ($\alpha$ = 0.08) & 50 & \textbf{2.28 $\scriptscriptstyle$$\pm$$\scriptscriptstyle$0.03} & \textbf{4.29 $\scriptscriptstyle$$\pm$$\scriptscriptstyle$0.02 } & \textbf{241.8 $\scriptscriptstyle$$\pm$$\scriptscriptstyle$ 0.9} & \textbf{336.37} & \textbf{7.62}\\
 FORA (n=2) & 50 & 2.65 $\scriptscriptstyle$$\pm$$\scriptscriptstyle$0.04 & 4.69 $\scriptscriptstyle$$\pm${$\scriptscriptstyle$0.03} & 238.5 $\scriptscriptstyle$$\pm$$\scriptscriptstyle$ 1.1 & 190.25 & 5.17 \\
  Ours ($\alpha$ = 0.18) &  50 & \textbf{2.65 $\scriptscriptstyle$$\pm$$\scriptscriptstyle$0.04} & \textbf{4.65 $\scriptscriptstyle$$\pm$$\scriptscriptstyle$0.03} & \textbf{238.7 $\scriptscriptstyle$$\pm$$\scriptscriptstyle$ 1.1}& \textbf{175.65} & \textbf{4.85}\\
 FORA (n=3) & 50 & 3.31 $\scriptscriptstyle$$\pm$$\scriptscriptstyle$0.05 & 5.71 $\scriptscriptstyle$$\pm${$\scriptscriptstyle$0.06} & 230.1 $\scriptscriptstyle$$\pm$$\scriptscriptstyle$ 1.3 & 131.81 & 4.12 \\
 Ours ($\alpha$ = 0.22) &  50 & \textbf{3.14 $\scriptscriptstyle$$\pm$$\scriptscriptstyle$0.05} & \textbf{5.19 $\scriptscriptstyle$$\pm$$\scriptscriptstyle$0.04} & \textbf{231.7 $\scriptscriptstyle$$\pm$$\scriptscriptstyle$ 1.0} & \textbf{131.81} & \textbf{4.11}\\
 \midrule
 No Cache  & 30 & 2.66 $\scriptscriptstyle$$\pm$$\scriptscriptstyle$0.04 & 4.42 $\scriptscriptstyle$$\pm${$\scriptscriptstyle$0.03} & 234.6 $\scriptscriptstyle$$\pm$$\scriptscriptstyle$ 1.0 & 219.36 & 4.88\\
 FORA (n=2) & 30 & 3.79 $\scriptscriptstyle$$\pm$$\scriptscriptstyle$0.04 & 5.72 $\scriptscriptstyle$$\pm${$\scriptscriptstyle$0.05} & 222.2 $\scriptscriptstyle$$\pm$$\scriptscriptstyle$ 1.2 & 117.08 & 3.13 \\
 Ours ($\alpha$ = 0.35) & 30 & \textbf{3.72 $\scriptscriptstyle$$\pm$$\scriptscriptstyle$0.04} & \textbf{5.51 $\scriptscriptstyle$$\pm$$\scriptscriptstyle$0.05}  & \textbf{222.9 $\scriptscriptstyle$$\pm$$\scriptscriptstyle$ 1.0} & \textbf{117.08} & \textbf{3.13}\\
 \midrule
  No Cache  & 70 & 2.17 $\scriptscriptstyle$$\pm$$\scriptscriptstyle$0.02 & 4.33 $\scriptscriptstyle$$\pm$$\scriptscriptstyle$0.02 & 242.3 $\scriptscriptstyle$$\pm$$\scriptscriptstyle$ 1.6 & 511.83 & 11.47\\
 FORA (n=2) & 70 & 2.36 $\scriptscriptstyle$$\pm$$\scriptscriptstyle$0.02 & 4.46 $\scriptscriptstyle$$\pm$$\scriptscriptstyle$0.03 & 242.2 $\scriptscriptstyle$$\pm$$\scriptscriptstyle$ 1.3 & 263.43 & 7.15 \\
 Ours ($\alpha$ = 0.08) & 70 & \textbf{2.37 $\scriptscriptstyle$$\pm$$\scriptscriptstyle$0.02} & \textbf{4.29 $\scriptscriptstyle$$\pm$$\scriptscriptstyle$0.03}  & \textbf{242.6 $\scriptscriptstyle$$\pm$$\scriptscriptstyle$ 1.5} & \textbf{248.8} & \textbf{6.9}\\
 FORA (n=3) & 70 & 2.80 $\scriptscriptstyle$$\pm$$\scriptscriptstyle$0.02 & 5.38 $\scriptscriptstyle$$\pm$$\scriptscriptstyle$0.04 & 238.0$\scriptscriptstyle$$\pm$$\scriptscriptstyle$ 1.2 & 175.77 & 5.61 \\
 Ours ($\alpha$ = 0.12) & 70 & \textbf{2.68 $\scriptscriptstyle$$\pm$$\scriptscriptstyle$0.02} & \textbf{4.90 $\scriptscriptstyle$$\pm$$\scriptscriptstyle$0.04}  & \textbf{238.8 $\scriptscriptstyle$$\pm$$\scriptscriptstyle$ 1.3} & \textbf{175.77} & \textbf{5.62}\\
\bottomrule
\end{tabularx}
\label{tab:dit_table}
\vspace{-6mm}
\end{table}

\begin{table}[b]
\centering
\vspace{-1mm}
\caption{Results For OpenSora on Rectified Flow (30 steps).}
\fontsize{5}{5.75}\selectfont
\setlength{\tabcolsep}{2.5pt}
\vspace{-1mm}
\begin{tabularx}{1\columnwidth}{c|c c c c| c c}
\toprule
Schedule & VBench (\%) ($\uparrow$) & LPIPS ($\downarrow$)  & PSNR ($\uparrow$)  & SSIM ($\uparrow$)  & TMACs & Latency (s) \\
\midrule
 No Cache &   79.36 $\scriptscriptstyle$$\pm$$\scriptscriptstyle$0.19& - & - & - & 1612.1 & 28.43\\
 Ours ($\alpha$ = 0.02) &   78.76 $\scriptscriptstyle$$\pm$$\scriptscriptstyle$0.38 & 0.5852 $\scriptscriptstyle$$\pm$$\scriptscriptstyle$ 0.0352 & 11.08 $\scriptscriptstyle$$\pm$$\scriptscriptstyle$ 0.96 & 0.5085 $\scriptscriptstyle$$\pm$$\scriptscriptstyle$ 0.0315 & 1388.5 & 26.57 \\
 Ours ($\alpha$ = 0.03) &   78.10 $\scriptscriptstyle$$\pm$$\scriptscriptstyle$0.51 & 0.5347 $\scriptscriptstyle$$\pm$$\scriptscriptstyle$ 0.1119 & 12.62 $\scriptscriptstyle$$\pm$$\scriptscriptstyle$ 2.81 & 0.5601 $\scriptscriptstyle$$\pm$$\scriptscriptstyle$ 0.0812  & 1321.1 & 26.17 \\
\bottomrule
\end{tabularx}
\label{tab:opensora_table}
\end{table}

\begin{table*}[h]
\centering
\vspace{-1mm}
\caption{Results For Stable Audio Open on DPMSolver++(3M) SDE on 3 datasets.}
\fontsize{6}{5}\selectfont
\renewcommand{\arraystretch}{1.7}
\setlength{\tabcolsep}{2.5pt}
\vspace{-1mm}

\begin{tabularx}{1.95\columnwidth}{c ccc  ccc  ccc  cc}
\toprule
& \multicolumn{3}{c}{AudioCaps} & \multicolumn{3}{c}{\shortstack{MusicCaps \\(No Singing)}} & \multicolumn{3}{c}{\shortstack{Song Describer \\(No Singing)}} & & \\
\cmidrule(lr){2-4} \cmidrule(lr){5-7} \cmidrule(lr){8-10}
Schedule & \texorpdfstring{FD\textsubscript{OpenL3}}{FDopenl3} ($\downarrow$) & \texorpdfstring{KL\textsubscript{PaSST}}{KLpasst} ($\downarrow$) & CLAP ($\uparrow$)& \texorpdfstring{FD\textsubscript{OpenL3}}{FDopenl3} ($\downarrow$) & \texorpdfstring{KL\textsubscript{PaSST}}{KLpasst} ($\downarrow$) & CLAP ($\uparrow$) & \texorpdfstring{FD\textsubscript{OpenL3}}{FDopenl3} ($\downarrow$) & \texorpdfstring{KL\textsubscript{PaSST}}{KLpasst} ($\downarrow$)& CLAP ($\uparrow$) & TMACs & Latency (s) \\
\midrule
No Cache & 81.7 $\scriptscriptstyle$$\pm$$\scriptscriptstyle$  6.8 & 2.13 $\scriptscriptstyle$$\pm$$\scriptscriptstyle$ 0.02  & 0.287 $\scriptscriptstyle$$\pm$$\scriptscriptstyle$ 0.003 & 
82.7 $\scriptscriptstyle$$\pm$$\scriptscriptstyle$ 2.1 & 0.931 $\scriptscriptstyle$$\pm$$\scriptscriptstyle$ 0.012 & 0.467$\scriptscriptstyle$$\pm$$\scriptscriptstyle$ 0.001 & 
105.2 $\scriptscriptstyle$$\pm$$\scriptscriptstyle$ 6.3 & 0.551 $\scriptscriptstyle$$\pm$$\scriptscriptstyle$0.024 & 0.421 $\scriptscriptstyle$$\pm$$\scriptscriptstyle$ 0.003 & 
209.82 & 5.65\\
Ours ($\alpha$ = 0.15) & 84.5 $\scriptscriptstyle$$\pm$$\scriptscriptstyle$ 6.7 & 2.15 $\scriptscriptstyle$$\pm$$\scriptscriptstyle$ 0.02 & 0.285 $\scriptscriptstyle$$\pm$$\scriptscriptstyle$ 0.003  &
85.9 $\scriptscriptstyle$$\pm$$\scriptscriptstyle$ 2.3 & 0.942 $\scriptscriptstyle$$\pm$$\scriptscriptstyle$ 0.012 & 0.467 $\scriptscriptstyle$$\pm$$\scriptscriptstyle$ 0.001 &
106.2 $\scriptscriptstyle$$\pm$$\scriptscriptstyle$ 6.6 & 0.555 $\scriptscriptstyle$$\pm$$\scriptscriptstyle$ 0.024 & 0.420 $\scriptscriptstyle$$\pm$$\scriptscriptstyle$ 0.003 &
170.75 & 4.59\\
Ours ($\alpha$ = 0.30) & 89.6 $\scriptscriptstyle$$\pm$$\scriptscriptstyle$ 6.3 & 2.17 $\scriptscriptstyle$$\pm$$\scriptscriptstyle$ 0.02 & 0.271 $\scriptscriptstyle$$\pm$$\scriptscriptstyle$ 0.003 &
82.0 $\scriptscriptstyle$$\pm$$\scriptscriptstyle$ 1.5 & 0.962 $\scriptscriptstyle$$\pm$$\scriptscriptstyle$ 0.012 & 0.448 $\scriptscriptstyle$$\pm$$\scriptscriptstyle$ 0.001 &
131.3 $\scriptscriptstyle$$\pm$$\scriptscriptstyle$ 5.9 & 0.596 $\scriptscriptstyle$$\pm$$\scriptscriptstyle$ 0.028 & 0.392 $\scriptscriptstyle$$\pm$$\scriptscriptstyle$ 0.003 &
136.16 & 3.72\\
\bottomrule
\end{tabularx}
\label{tab:stableaudio_table}
\vspace{-4mm}
\end{table*}


\begin{figure*}[h]
  \centering
  \includegraphics[width=0.9\textwidth]{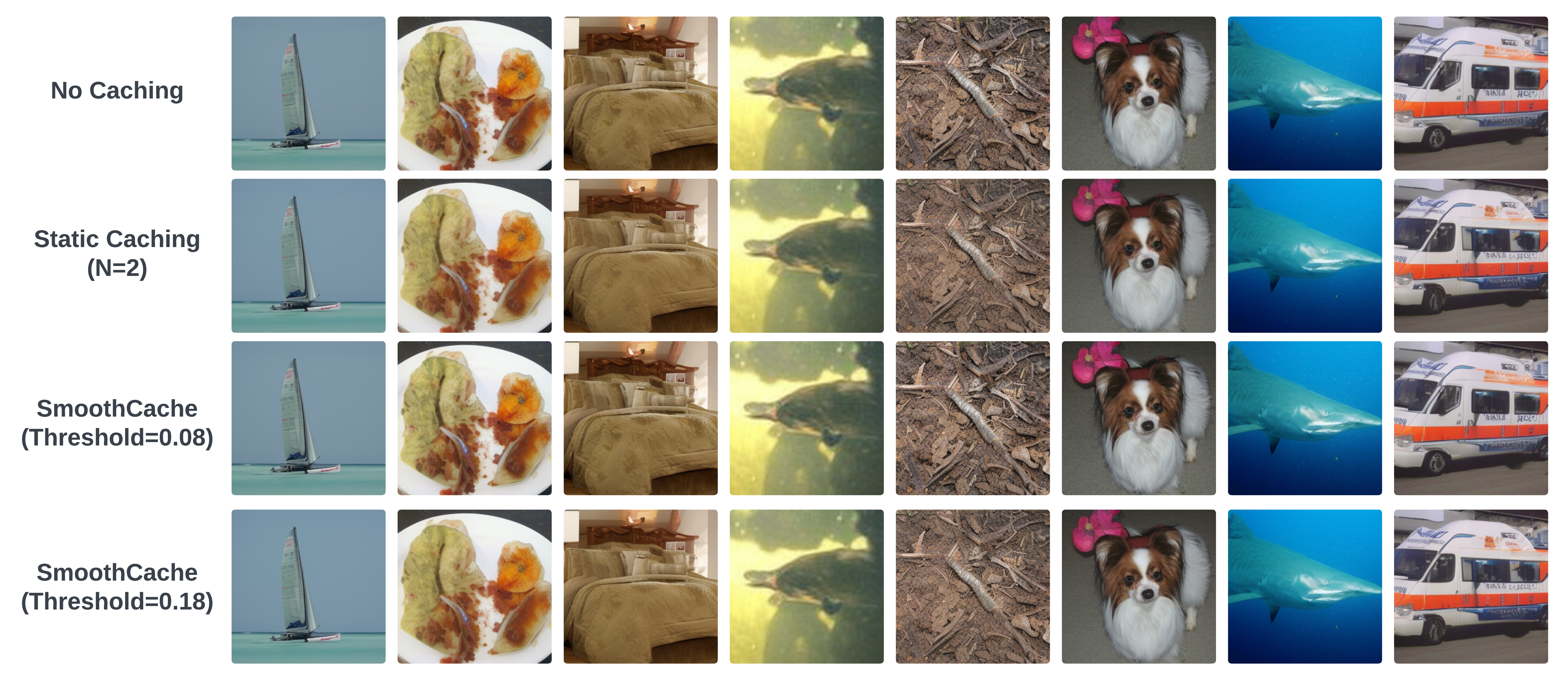}
\vspace{-4mm}
  \caption{\textbf{SmoothCache results on DiT-XL/2-256$\times$256} for unconditional generation with 50 DDIM sampling steps on ImageNet-1k for thresholds 0.08 and 0.18, as well as for Static Caching.}
  \label{fig:imagenet1k_qual}
\vspace{-6mm}
\end{figure*}
\begin{figure*}[t]
  \centering
  \includegraphics[width=0.85\textwidth]{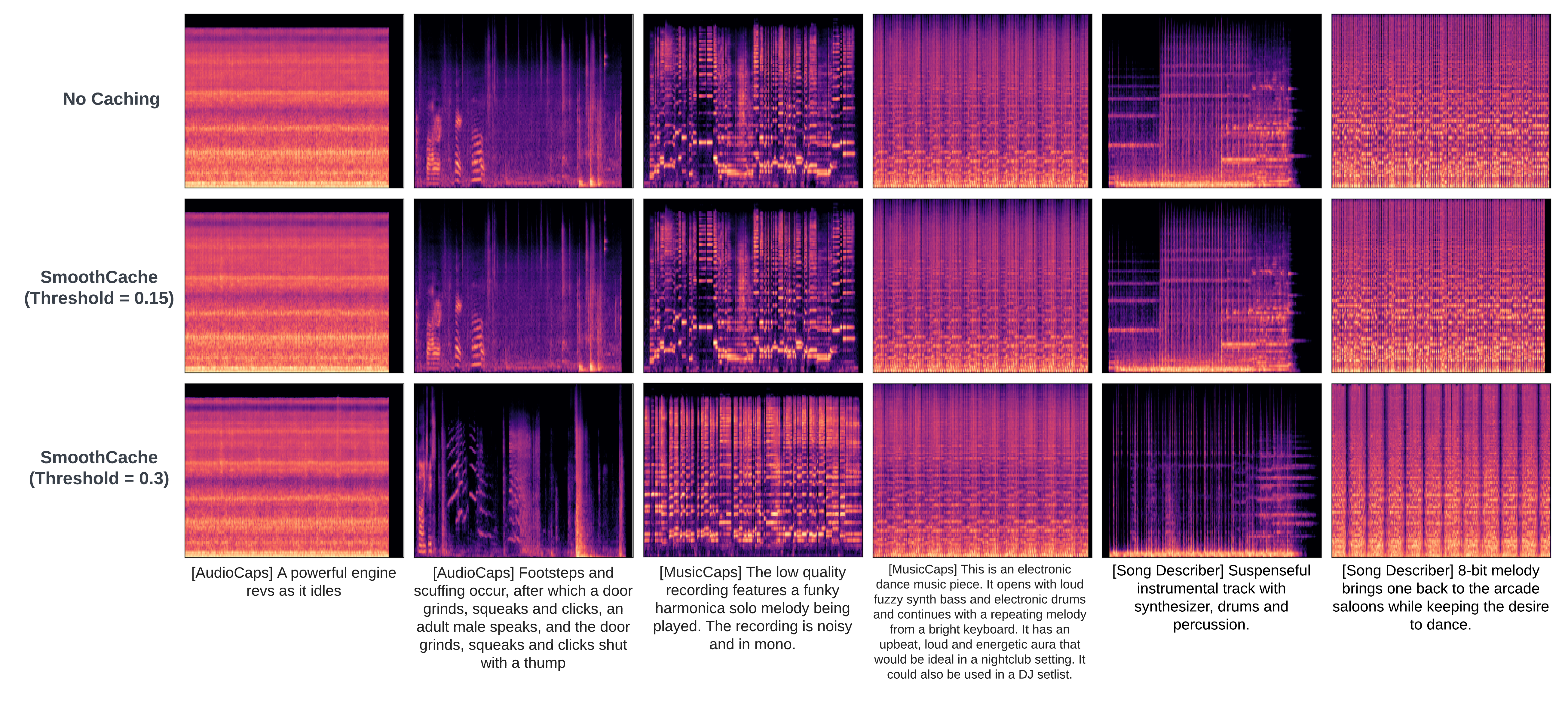}
\vspace{-5mm}
  \caption{\textbf{SmoothCache Results on Stable Audio Open} for threshold 0.15 and 0.3. Log-Mel Spectrograms are shown.}
  \label{fig:stable_audio_qual}
\vspace{-7mm}
\end{figure*}
\begin{figure}[b]
\vspace{-6mm}
  \centering
  \includegraphics[width=0.85\columnwidth]{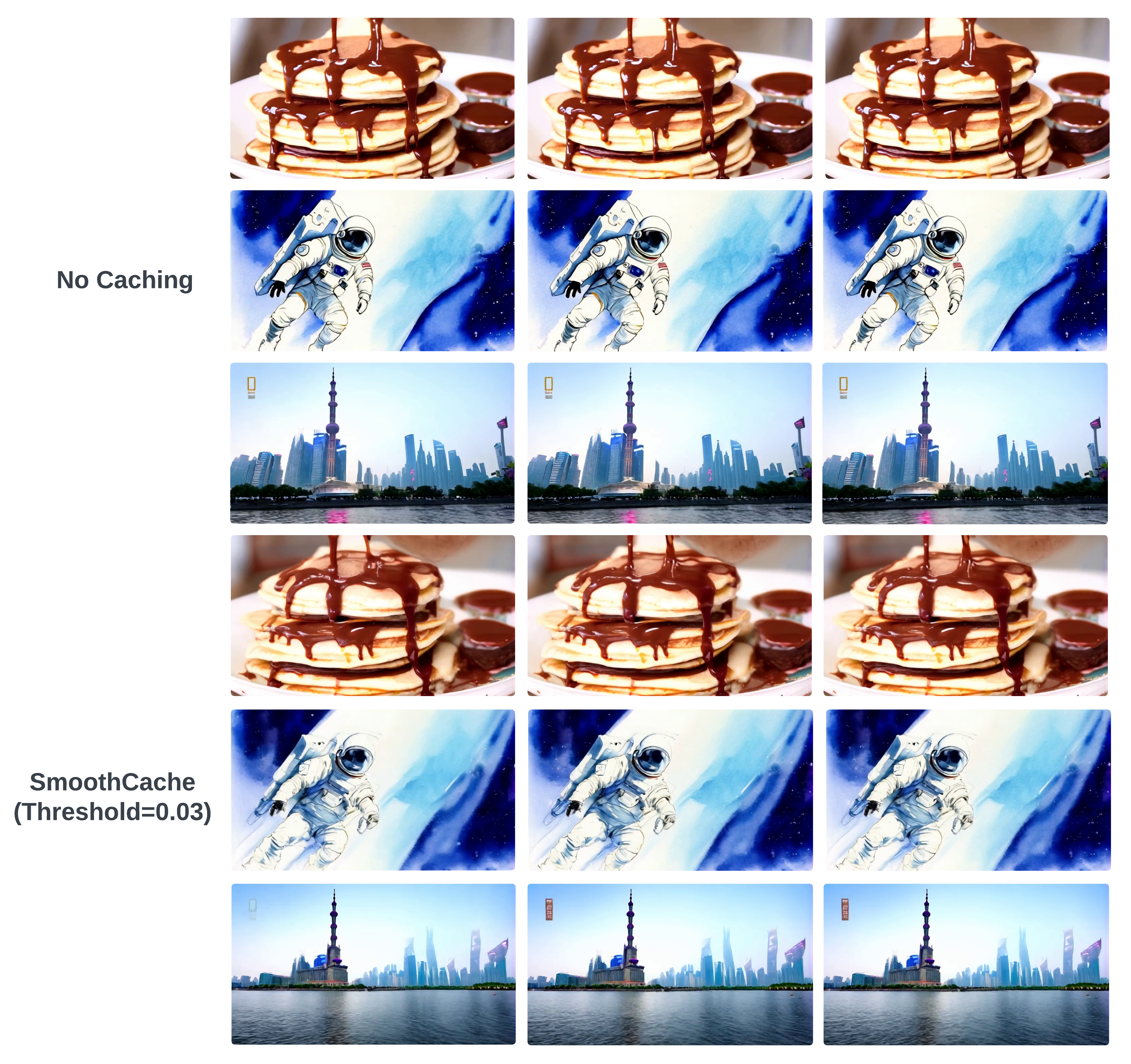}
  \caption{\footnotesize\textbf{SmoothCache Results on OpenSora} for threshold 0.03 for 2s 480p videos. We show the first, middle and last frame of each video. We use the following prompts, in order from top to bottom: (1) \texttt{Chocolate sauce is poured slowly over a stack of fluffy pancakes.} (2) \texttt{an astronaut floating in space with the earth in the background} (3) \texttt{The bund Shanghai, pan right} }
  \label{fig:opensora_qual}
\end{figure}

\subsubsection{Quantitative results}

\textbf{Comparing against existing literature}
In order to further investigate the true effectiveness of our technique, we compare SmoothCache to existing DiT caching techniques in literature developed concurrently with SmoothCache, such Fast-Forward Caching (FORA)~\cite{selvaraju2024forafastforwardcachingdiffusion} and Learning-to-Cache (L2C)~\cite{ma2024learningtocacheacceleratingdiffusiontransformer} for Label-to-Image generation. We note that the above methods relies on specific properties of the relevant model architecture, and does not necessarily translate across different modalities, while SmoothCache is agnostic to those considerations as it models the caching scheme directly off the observed error curves. FORA does not work in OpenSora or Stable Audio Open due to the difference in the error curves as seen in Fig.~\ref{fig:component_error_curves} and hence we do not report the results here. We see that SmoothCache outperforms FORA across different inference times at 50 sampling steps, yielding similar performance at lower inference time or better performance with the same inference time. The one exception  here is L2C, which requires leveraging the full ImageNet training set to learn a policy for a given number of sampling steps. Changing the number of sampling steps requires a full retraining, and L2C has a theoretical maximum of a 2$\times$ speedup because the caching policy is only learned with skipping every other step, limitations that SmoothCache does not have.

\textbf{Examining inference speedup/quality tradeoff} Stable Audio Open and DiT-XL/2-256$\times$256 provides the highest speed/quality tradeoff compared to OpenSora, which gives around a 10\% speedup latency wise and around 16-22\% reduction in MACs comapred to an almost 20-60\% speedup for the other modalities, whose discrepancy could be attributed by the larger overhead of non-DiT components (since inference latency is measured end-to-end).
We also observe that the amount of inference speed/quality tradeoff looks directly correlated to the error deviation between calibration samples which can be seen in Fig.~\ref{fig:component_error_curves}, with the higher variance in error among individual samples across timesteps for OpenSora versus that for Stable Audio Open and DiT-XL/2-256$\times$256.

\subsubsection{Qualitative results}
We show visual examples for all modalities. We show 256$\times$256 images for image generation, using null-conditional prompts. For audio, we show the log-mel spectrogram in order to visualize the waveform for conditional prompt generation across the 3 datasets we measure. For video, we show the first, middle and last frame of 2 second 480p 24 fps videos at a 9:16 aspect ratio.

We see that for DiT-XL/2-256$\times$256 outputs in Fig.~\ref{fig:imagenet1k_qual}, even though there is a drop of FID with SmoothCache applied, that there is some noticeable difference in performance quality when there is a higher threshold applied, but there exists a threshold where the quality of the model performance is visually indistinguishable from the model without caching. This shows the fine-granularity in inference speed/quality tradeoff that SmoothCache affords, while static caching as shown in FORA can only yield the lower quality SmoothCache threshold.
We observe particularly that this degradation is less pronounced in Stable Audio Open in Fig.~\ref{fig:stable_audio_qual}, where the audio waveforms look visually identical, and when listened to do not have perceptible differences. This is consistent with the qualitative results which show only a minor degradation across all measured metrics across all datasets.
For OpenSora, we see more significant differences in when caching in Fig.~\ref{fig:opensora_qual}, with noticeable artifacting when the higher caching threshold is applied. This is consistent with the quantitative results, and suggests this architecture/modality is more sensitive to caching than the other modalities, which makes sense due to the complex spatial/temporal modelling that other modalities explored do not have to deal with.


\subsection{Ablations}
\label{ablations}
We ablate our investigations by showing how model performance varies across different step sizes, and show the robustness of SmoothCache for the DiT-XL-256$\times$256 model.

\textbf{Examining the Caching/Sample Step Pareto Front}
We attempt to show via comparison with static caching that our technique yields a better Pareto front along different sampling steps through the image generation results. We note that all caching techniques have varying performance as you vary the number of solver sampling steps, and we show that despite to the minimal assumptions SmoothCache makes about the sampler and model architecture, SmoothCache at least matches up to other caching strategies if not outright beating them across multiple fronts. Table~\ref{tab:dit_table} shows that even with lower or higher number of sampling steps, SmoothCache outperforms FORA across multiple inference speed/quality points at different number of sampling steps. 
We observe that SmoothCache outperforms Static Caching for multiple sampling configurations. While L2C slightly outperforms SmoothCache on DDIM sampling for DiT-XL, SmoothCache does not require expensive training over ImageNet1k, and generalizes to different model architectures and sampling configurations.

\textbf{Calibration sample size}
We note that the choice of the number of calibration samples does not adversely affect the caching schedule generated. Empirically, we observe that 10 samples for all 3 models investigated in this paper is usually enough to reliably regenerate the same caching schedule given the same $\alpha$. We however note that the confidence interval of the different error curves vary from modality to modality as seen in Fig.~\ref{fig:component_error_curves}.

\section{Limitations and Future Work}
\label{limitation_future_work}

The main limitation of the SmoothCache technique is its reliance on the repeated DiT block architecture, particularly the residual connections following the aforementioned computational bottleneck layers. These connections allow for the output $y_t$ to be approximated by some linear function of input and previously cached layers $f(x_{t+k}) + x_t$, a feature we consider essential for its caching effectiveness. Additionally, the performance gains from SmoothCache are strongly linked to the computational intensiveness of the candidate layers. Consequently, it may yield smaller performance improvements when applied to DiT networks with limited depth or width.

A secondary limitation lies in the assumption that errors from approximating outputs of earlier layers have minimal impact on the loss function guiding caching decisions for deeper layers. For instance, if the first Self-attention layer is approximated using a cached result from a previous timestep, caching the second Self-attention layer could introduce further output discrepancies, even if the calibration loss suggests low error. This is because the calibration loss is computed when no caching is performed, which may not fully model true approximation errors during SmoothCache-enabled inference. We address this issue by grouping caching and computation decisions for layers of the same type at each timestep. However, this does not fully resolve dependency issues between different layer types, leaving room for further optimization in future work. 

We also highlight a phenomenon that future work can investigate where the pareto front of inference speed/quality seems to correlate with the variance in error across different calibration samples, with architectures/modalities that have higher variance between sample error curves having narrower fronts than those which have lower variance.

%% file: sec/3_conclusion.tex
\section{Conclusion}
We introduce SmoothCache, a training-free caching technique that works across multiple diffusion solvers and modalities.  Using the layer representation error of a calibration inference pass, SmoothCache identifies redundancies in the diffusion process, enabling caching and reuse of output feature maps, reducing the number of computationally expensive layer operations. 
Our evaluations show it at least matches or exceeds the performance of existing modality-specific caching methods.

%% file: sec/X_suppl.tex
\clearpage
\setcounter{page}{1}

\twocolumn[{%
\renewcommand\twocolumn[1][]{#1}%
\maketitlesupplementary
\begin{center}
    \vspace{-4mm}
    \centering
    \captionsetup{type=figure}
    \includegraphics[width=0.8\textwidth]{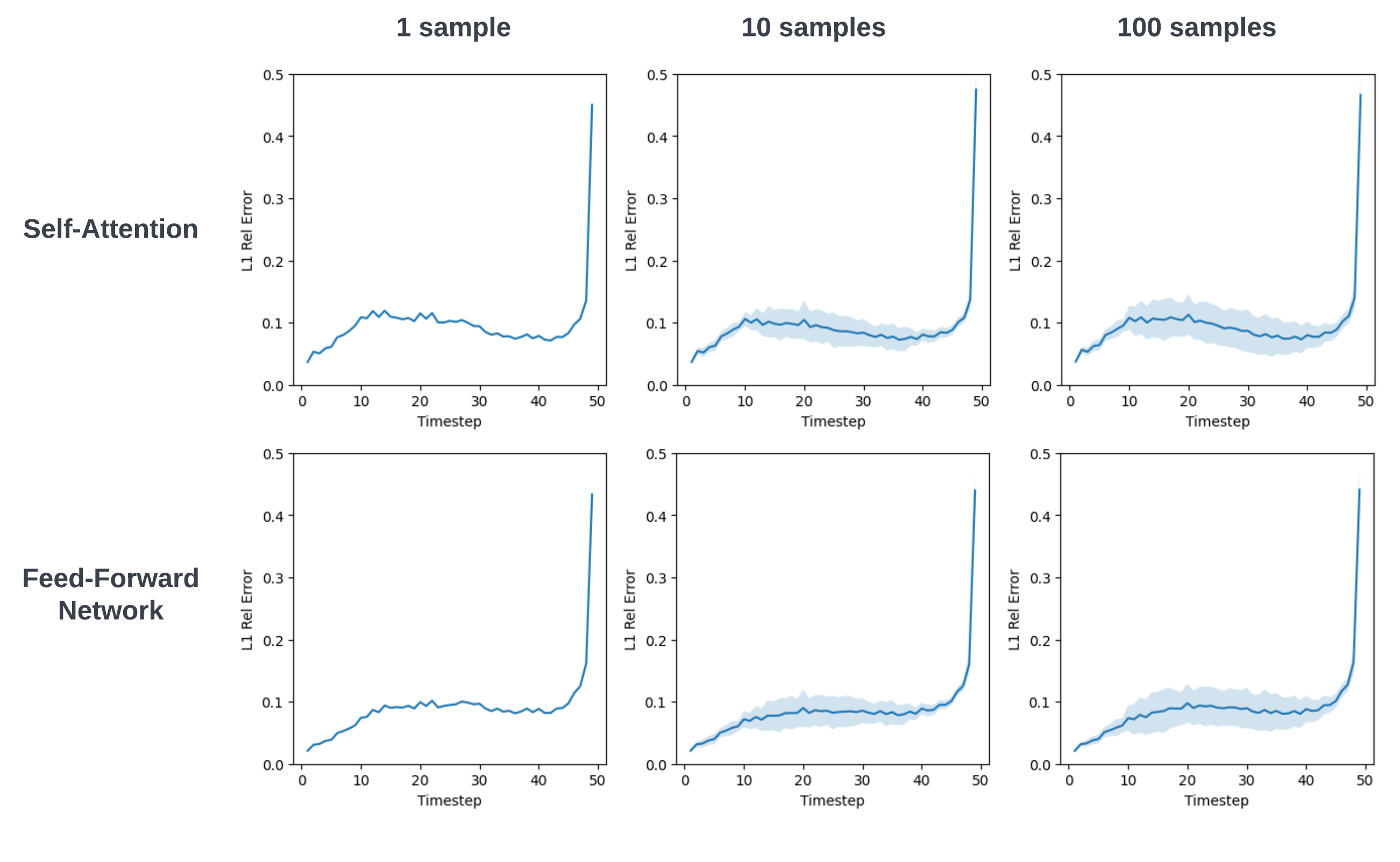}
    \vspace{-8mm}
    \caption{\textbf{L1 Relative Error Curves of different architecture components for DiT-XL}. Curves are plotted with 95\% confidence intervals and scaled to the same y-axis range.}
    \label{sec:supp_calib_samples}
\end{center}%
}]

\section{Ablations Addendum: Number of Calibration Samples}
As mentioned in Sec. \ref{ablations}, the calibration sample size chosen does not dramatically change the generated schedule. Increasing the number of samples only affects the range of the confidence interval for the error curves, but not the mean. Future methods that might leverage the uneven distribution of error curves between samples might need to take note of this observation.